\pdfoutput=1

\documentclass[11pt]{article}

\usepackage[]{EMNLP2022}

\usepackage{times}
\usepackage{latexsym}

\usepackage[T1]{fontenc}

\usepackage[utf8]{inputenc}

\usepackage{microtype}

\usepackage{inconsolata}

\usepackage{hyperref}
 \usepackage{geometry}
 \usepackage{tikz}
\usepackage{comment}
\usepackage{amsmath,amssymb} 
\usepackage{color}
\usepackage{graphicx}
\usepackage{booktabs}
\usepackage{array}
\usepackage{relsize}
\usepackage{multirow}
\usepackage{caption}
\usepackage{subcaption}
\usepackage{tabularx}
\usepackage{floatrow}

\usepackage[capitalize]{cleveref}
\crefname{section}{Sec.}{Secs.}
\Crefname{section}{Section}{Sections}
\Crefname{table}{Table}{Tables}
\crefname{table}{Tab.}{Tabs.}

\renewcommand{\comment}[1]{}
\newcommand{\vv}{\boldsymbol{v}}
\newcommand{\textset}{\mathcal{T}}
\newcommand{\clipim}[1]{\boldsymbol{\phi}(#1)}
\newcommand{\cliptx}[1]{\boldsymbol{\psi}(#1)}
\newcommand{\nv}{\boldsymbol{n}}
\newcommand{\reals}{\mathbb{R}}
\newcommand{\modelname}{CapDec}

\title{Text-Only Training for Image Captioning using Noise-Injected CLIP}

\author{
        David Nukrai 
        \And Ron Mokady \\ Blavatnik School of Computer Science, Tel Aviv University 
        \And Amir Globerson 
        }

\begin{document}
\maketitle
\begin{abstract}
We consider the task of image-captioning using only the CLIP model and additional text data at training time, and no additional captioned images. 
Our approach relies on the fact that CLIP is trained to make visual and textual embeddings similar. 
Therefore, we only need to learn how to translate CLIP textual embeddings back into text, and we can learn how to do this by learning a decoder for the frozen CLIP text encoder using only text. We argue that this intuition is ``almost correct'' because of a gap between the embedding spaces, and propose to rectify this via noise injection during training.
We demonstrate the effectiveness of our approach by showing SOTA zero-shot image captioning across four benchmarks, including style transfer. 
Code, data, and models are available at \url{https://github.com/DavidHuji/CapDec}.

\end{abstract}

\begin{figure*}

    \includegraphics[width=0.87\textwidth]{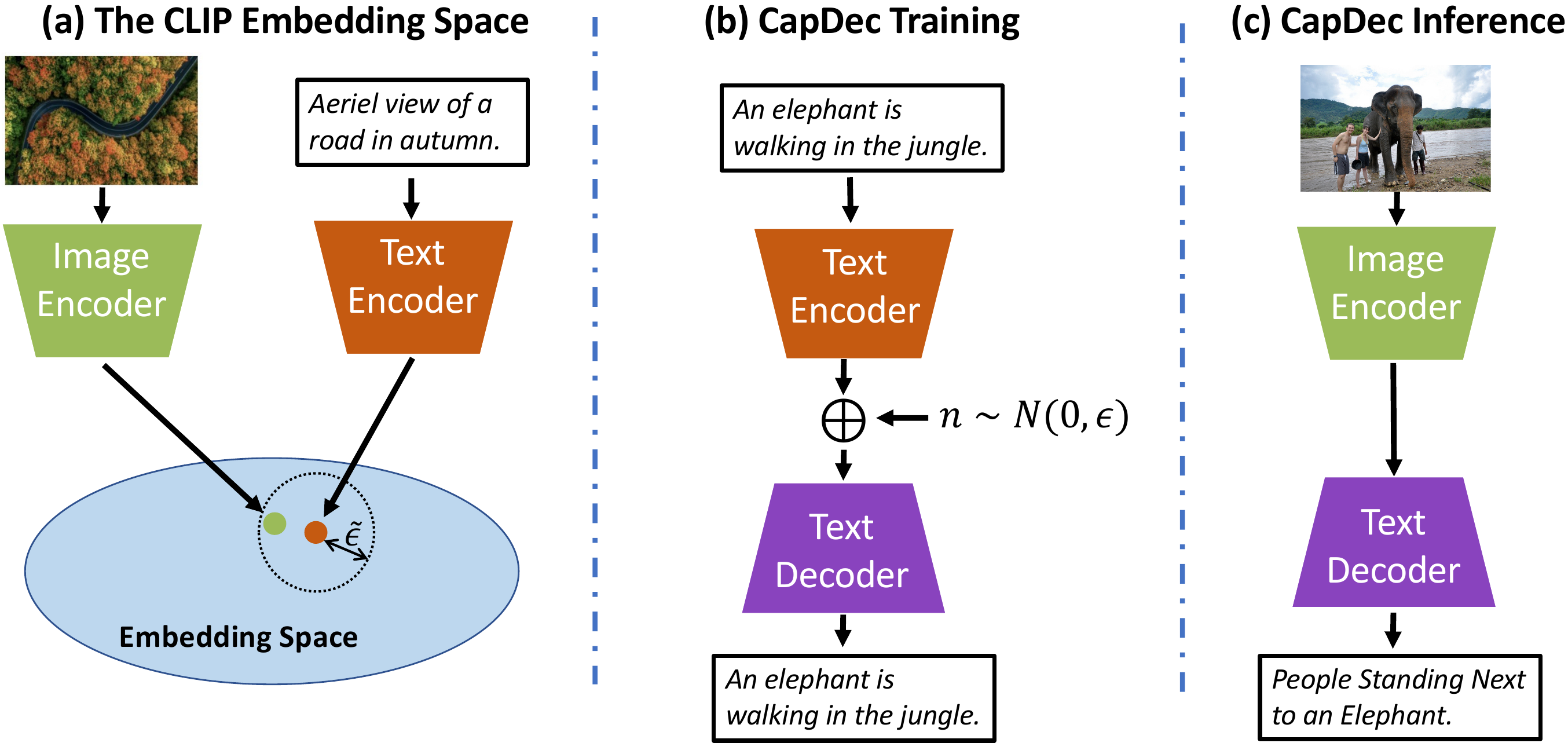}
    \caption{\textbf{Overview of our \modelname{} captioning approach. (a)} An illustration of the CLIP joint embedding space. Embedded text is relatively close to its corresponding visual embedding, but with a certain gap. \textbf{(b)} \modelname{} trains a model that decodes the CLIP embedding of text $T $ back to text $T$, after noise-injection. The encoders remain frozen. \textbf{(c)} At inference, \modelname{} simply decodes the embedding of an image using the trained decoder.}
    \vspace{-0.15cm}
    \label{fig:method_diagram}
\end{figure*}

\section{Introduction}

Vision and language are closely intertwined, as they are two ways of describing the world. This raises the potential for developing models that map images and text into a shared semantic space. Indeed, this approach has recently achieved great success with models like CLIP \cite{radford2021learning} and ALIGN \cite{jia2021scaling}. These models use \textit{parallel} image-text data to train a joint representation, where the embeddings of image-text pairs are similar. Such models have been employed for various vision-language tasks. 



Image captioning is a key task in vision-language perception. Yet,
training image captioning models typically requires large datasets of captioned images, and these are challenging to collect. Furthermore, it is not clear how one could adapt a pretrained vision-language model to generate captions in new styles. In this work, we present an approach to captioning that only requires CLIP and text data, and generates styled captions using only unpaired textual examples from that style. This alleviates the need for paired text-image data, and also allows for simple style transfer.


A first approach one could consider for this setting is to train a decoder model to reconstruct texts from their respective CLIP embeddings, and at inference use this decoder to decode image embeddings. However, we observed that this approach fails at inference, and we conjecture this is due to the known domain gap between the image and text modalities \cite{liang2022mind}. We propose a simple approach to mitigate this, by injecting noise into the embedding during training. This has the effect of creating a ball in embedding space that will map to the same caption, and corresponding image-embedding is more likely to be inside this ball, as illustrated in \cref{fig:method_diagram}.a.

We evaluate our ``Captioning via Decoding'' (CapDec) method extensively, showing that it  works well on several image captioning tasks, including standard, cross-domain, and style-guided captioning.
Overall, our main contributions are as follows:
1) 
A simple and intuitive approach to learning a captioning model based on CLIP and additional text training data, but no images for training. 
2) Evaluation of \modelname{} on image captioning tasks, including generating captions in various styles, shows it outperforms other methods which use the same supervision.


\comment{
Vision and language are closely intertwined, as they are two ways of describing the world. This raises the potential of developing models that map images and text into a shared semantic space. Indeed, this approach has recently achieved great success with models like CLIP \cite{radford2021learning} and ALIGN \cite{jia2021scaling}, which uses parallel image-text data to train a joint representation of image and text.

Specifically, such models learn two embedding functions: a visual-embedding from images to $\reals^d$ and a text-embedding from texts to $\reals^d$. These embeddings are trained such that a for an image $I$ described by text $T$, the embeddings of $I$ and $T$ will be similar \amirg{in dot-product}.

One striking application of these embedddings is so-called zero-shot learning. For example, given an image $I$ containing a cat or a dog, we can check whether its visual-embedding is closer to the text-embedding of ``This is a photo of a cat'' or ``This is a photo of a dog'', and classify accordingly. Importantly, this can be done without training on any labeled images. 

While the above works well for categorization, it is challenging to extend it to the problem of caption generation, where we seek a textual description of an input image. Presumably, we can search over all possible textual descriptions and find the one closest to the image in embedding space. However, this will not work for two reasons. First, it is computationally hard to consider all possible captions. Second, as we show later, the textual embedding space is non-robust in the sense that there are texts that will be mapped close to an image, but will not describe it well. 

Here, we provide an approach that overcomes the above two challenges. To overcome the computational challenge, we train a decoder from CLIP embedding space to text. To train it, we take an autoencoding approach: we simply take sentences $s$, encode them via CLIP, and train the decoder to output $s$. However, this still leaves the problem of non-robustness of the embeddings. Namely, small changes to embedding space result in large changes in semantics of the ouptut sentence
To overcome this, we inject noise after encoding, so that the decoder will be less sensitive to noise.

We show that our approach works well on several tasks requiring captions: standard image-free captioning, cross-domain captioning and caption style transfer \amirg{complete}.
}

\section{Related Work}
\label{sec:rw}

Image captioning methods \cite{chen2014learning,  chen2017sca,yang2019learning, herdade2019image, luo2021dual, tsimpoukelli2021multimodal} typically extract visual features using a pre-trained network. These are passed to a textual decoder that produces the final captions. To bridge the gap between vision and language, other works employ pre-training to create a shared latent space of vision and text \cite{tan2019lxmert, laina2019towards, lu2019vilbert, li2020oscar, zhou2020unified, zhang2021vinvl, wang2021simvlm, hu2022scaling}. However, all of these approaches require extensive training and large paired datasets that are hard to collect. Gan et al.~\shortcite{gan2017stylenet} and Zhao et al.~\shortcite{zhao2020memcap} have suggested style-guided captioning, but also employ training over paired data.

CLIP \shortcite{radford2021learning} marked a turning point in vision-language perception, and has been utilized for vision-related tasks by various distillation techniques \cite{gu2021open, song2022clip, jin2021object, gal2021stylegan, xu2021simple, khandelwal2022simple}.
Recent captioning methods use CLIP for reducing training time \cite{mokady2021clipcap}, improved captions \cite{shen2021much, Luo2022AFS, luo2022tuning, Cornia2021UniversalCL, kuo2022beyond}, and in zero-shot settings \cite{su2022language, tewel2022zerocap}. However, zero-shot techniques often result in inferior performance, as the produced captions are not compatible with the desired target style, which is usually dictated by a dataset.
In this work, we suggest a new setting, where we adapt CLIP to image captioning using only textual data. 
As a result, we can easily adapt captions to any desired caption style given instances of text in that style. 
Concurrent work by \citet{su2022language} efficiently produces high-quality captions with the minimal supervision of text-only pre-training by employing CLIP-induced score at inference. Our approach is arguably simpler and also outperforms \citet{su2022language} empirically. Note that \citet{zhou2021lafite} have also employed noise-injection, but for
the opposite problem of CLIP-based text-free text-to-image generation.

\begin{table*}
{ \small
$(A)$ \textbf{Image Captioning}  \\
\begin{tabular}{lcccccccccc}
    \toprule
    \multirow{3}{50pt}{\textbf{Model}} &
    \multicolumn{5}{c}{MS-COCO} &
    \multicolumn{5}{c}{Flickr30k}  \\
    \cmidrule(lr){2-6} \cmidrule(lr){7-11} &
    B@1 & B@4 & M & R-L & CIDEr & 
    B@1 & B@4 & M & R-L & CIDEr \\
    \midrule
    \multicolumn{11}{c}{\textit{Fully Supervised Approaches}} \\
    \midrule
    BUTD & 77.2 & 36.2 & 27.0 & 56.4 & 113.5 & - & 27.3 & 21.7 & - & 56.6  \\
    UniVLP & - & 36.5 & 28.4 & - & 116.9 & - & 30.1 & 23.0 & - & 67.4  \\
    ClipCap & 74.7 & 33.5 & 27.5 & - & 113.1 & - & 21.7 & 22.1 & 47.3 & 53.5  \\
    Oscar & - & 36.5 & 30.3 & - & 123.7 & - & - & - & - & - \\
    LEMON & - & 40.3 & 30.2 & - & 133.3 & - & - & - & - & - \\
    \midrule
    \multicolumn{11}{c}{\textit{Weakly or Unsupervised Approaches}} \\
    \midrule
    ZeroCap & 49.8 & 7.0 & 15.4 & 31.8 & 34.5 & 44.7 & 5.4 & 11.8 & 27.3 & 16.8  \\
    MAGIC & 56.8 & 12.9 & 17.4 & 39.9 & 49.3 & 44.5 & 6.4 & 13.1 & 31.6 & 20.4 \\
    
    \midrule
    
    \textbf{\modelname{}} & \textbf{69.2} & \textbf{26.4} & \textbf{25.1} & \textbf{51.8} & \textbf{91.8} & \textbf{55.5} & \textbf{17.7} & \textbf{20.0} & \textbf{43.9} & \textbf{39.1} \\
    
    \toprule
\end{tabular}
\vspace{0.15cm}

$(B)$ \textbf{Cross-Domain Captioning}
\begin{tabular}{lccccccccccc}
    \toprule
    \multirow{2}{50pt}{} &
    \multicolumn{5}{c}{Flickr30k $\Longrightarrow$ MS-COCO} &  
    \multicolumn{5}{c}{MS-COCO $\Longrightarrow$ Flickr30k}\\
    \cmidrule(lr){2-6} \cmidrule(lr){7-11} &
    B@1 & B@4 & M & R-L & CIDEr & 
    B@1 & B@4 & M & R-L & CIDEr  \\
    \midrule
    MAGIC & 41.4 & 5.2 & 12.5 & 30.7 & 18.3 & 46.4 & 6.2 & 12.2 & 31.3 & 17.5 \\
    \midrule
    
    \textbf{\modelname{}} & \textbf{43.3} & \textbf{9.2} & \textbf{16.3} & \textbf{36.7} & \textbf{27.3} & \textbf{60.2} & \textbf{17.3} & \textbf{18.6} & \textbf{42.7} & \textbf{35.7}\\
    
    \toprule
\end{tabular}


\caption{\textbf{Results for image captioning. (A)} We use captions from the COCO and Flickr30k to train \modelname{} and evaluate on the datasets the captions were taken from. We report results for fully supervised methods that train on captioned images, and on methods that use no training text (ZeroCap), or just training text and no images (\modelname{} and MAGIC). \textbf{(B)} Similar setting to (A), but in cross-domain setup where training text is taken from one dataset, and evaluation is done on the second dataset.
}
\vspace{-0.35cm}
\label{tab:captioning} 
}
\end{table*}

\section{Method}
\label{sec:method} 

\paragraph{Text-Only Training.}
Our goal is to learn a model that produces a caption for a given image $I$.
Unlike supervised approaches, we assume that during training we only have access to a set of texts $\textset$. These can be obtained by harvesting a text corpus. 
We next introduce notation for the CLIP model. Given an image $I$ let 
$\clipim{I}\in\reals^d$ be its embedding, and given a text $T$ let
$\cliptx{T}\in\reals^d$ be its embedding. For converting a vector $\vv\in\reals^d$ into a caption, we use a textual decoder $C(\vv)$ consisting of a lightweight mapping network and a pretrained auto-regressive language model, as suggested in \citet{mokady2021clipcap}.

We train the decoder as follows (except for the noise-injection which we introduce below). Each text $T\in\textset$ is first mapped to CLIP space via $\cliptx{T}$ and then decoded back into a text via $C(\cliptx{T})$. We would like this decoding to be similar to the original text $T$. Namely, our training objective is a reconstruction of the input text from CLIP's textual embedding.
At inference, given an image $I$ we simply apply the decoder to $\clipim{I}$, returning the caption $C(\clipim{I})$.

\paragraph{Noise-Injected CLIP Embeddings.}
We observed that the above training scheme results in inaccurate captions during inference. We conjecture this is because the embeddings of the text and image modalities are separated by a domain gap, as shown in \citet{liang2022mind}. As a result, while text reconstruction is successful during training, inference fails when using image embeddings instead. If image-text pairs were available, we could attempt to learn a mapping between these domains. Nevertheless, as we aim for text-only training, we shall seek a different approach. 

Specifically, we assume that the visual embedding corresponding to a text embedding lies somewhere within a ball of small radius $\epsilon$ around the text embedding (see \cref{fig:method_diagram}). We would like all text embeddings in this ball to decode to the same caption, which should also correspond to the visual content mapped to this ball. We implement this intuition by adding zero-mean Gaussian noise of STD $\epsilon$ to the text embedding before decoding it. 

The value of $\epsilon$ is calculated by estimating the spread of captions corresponding to the same image. Specifically, we set $\epsilon$ to the mean $\ell_{\infty}$ norm of embedding differences between five captions that correspond to the same image. We estimated this based on captions of only $15$ MS-COCO images. Since this calculation requires very few captions and there is no need to recalculate it for every new dataset, we do not view it as additional supervision.


Our overall training objective is thus to minimize:
\begin{equation}
\sum_{T\in\textset} \ell(C(\cliptx{T} + \nv), T) ~,
\end{equation}
where $\nv\in\reals^d$ is a random standard Gaussian noise with
standard-deviation $\epsilon$ and $\ell$ is an auto-regressive cross-entropy loss for all tokens in $T$. We train just the parameters of the textual decoder $C$, while the encoder $\cliptx{}$ is kept frozen. The noise is sampled independently at each application of the encoder. 




\comment{
\subsection{Generic ClipL2V Framework}

As illustrated in Figure 2 (TBD add a small figure, near the abstract, that illustrates our generic framework), we discovered that using noise regularization, it is possible to learn robust decoding from Clip’s language embeddings that generalizes well to Clip’s vision embeddings decoding. 

Our generic problem setting is as follows: given a vision task, vTask(Img) that we are also able to define a dual language task lTask(txt) s.t. for any pair <img, txt> we have task(txt) = task(img) (hypothetically if we had pairs). Assuming we have unpaired data of corpus <T>, and images set <Imgs>, our problem then is to learn a model M, s.t. M(img) $\sim$= vTask(img) without having access to the Imgs set in training.

For solving that, we suggest a novel image-free-training procedure for a model M that is trained only on the language domain while it is still robust for inference in the vision domain. I.e. Our training objective is $min_w(loss( M(text)), lTask(text) )$, while inference by: M(Img) --> vTask(Img). 
As illustrated in figure 2. (TBD add figure) our simple generic model is 
$M(sample) = robustDecoder(frozenClipEncoder(sample))$. 
The frozen clip encoder during training is clip’s text decoder while in inference it is its vision encoder.

\subsection{Image-Free Training}

We train M only on the text corpus by minimizing reconstruction loss: loss( M(text), text ), while we expect it to generalize to the vision domain. This is not a trivial expectation since the Clip’s language embeddings are not identical to Clip’s vision embeddings. We actually found that for image captioning it happens only partially. In order to mitigate the gap between the text and vision embeddings, we suggest a simple, yet effective regularization as follows. During training, we add gaussian noise to the clips’ embeddings, thus we enforce robust decoding. Specifically for every epoch and sample, we sample $x$$\sim$$N(0,Var)$, and we train by minimizing: $loss(text,  robustDecoder(frozenClipEncoder(sample) + x) )$.
Besides captioning, the suggested generic approach may be applied to many vision tasks such as classification, VQA, OD, referring expression, etc.  

\subsection{Image Captioning Model}
In this section we detail adaptation of the generic ClipL2V to the task of zero-shot image captioning. Our vTask is defined as $vTask(img)->caption$. Correspondingly, the language task is defined as $lTask(caption)->caption$. Inspired by ClipCap [ref] we use $robustDecoder(e)=GPT2(transformer(e))$. Figure 1. Demonstrates the full pipe line of our ClipL2V for image-free captioning. To sum up, our model is trained with text only, by the objective of $CE(text, GPT2(mapper(clipTextEncoder(text))))$, while in inference we use the clip vision encoder:  $GPT2(mapper(clipImageEncoder(image)))$. 
}

\begin{table*}
{ \small
\begin{tabular}{lcccccccc}
    \toprule
    \multirow{2}{50pt}{\textbf{Model}} &
    \multicolumn{4}{c}{Romantic} & 
    \multicolumn{4}{c}{Humorous} \\
    \cmidrule(lr){2-5} \cmidrule(lr){6-9} &
    B@1 & B@3 & M & C & B@1 & B@3 & M & C \\
    \midrule
    StyleNet & 13.3 & 1.5 & 4.5 & 7.2 & 13.4 & 0.9 & 4.3 & 11.3 \\
    \midrule
    MemCap & 21.2 & 4.8 & 8.4 & 22.4 & 19.9 & 4.3 & 7.4 & 19.4 \\
    \midrule
    \modelname{} + Image-Text Pre-training & 27.9 & 8.9 & 12.6 & 52.2 & 29.4 & 8.8 & 13.2 & 55.1 \\
    \midrule
    \modelname{} + Text-Only Pre-training & 23.0 & 4.6 & 9.1 & 27.4 & 22.7 & 4.3 & 9.7 & 29.0 \\
    \midrule
    \modelname{}  & 21.4 & 5.0 & 9.6 & 26.9 & 24.9 & 6.0 & 10.2 & 34.1 \\
    \toprule
\end{tabular}
\caption{\textbf{Style-Guided captioning results on FlickrStyle10K }\cite{gan2017stylenet}. 
}
\vspace{-0.35cm}
\label{tab:style} 
}
\end{table*}
\section{Results}
\label{sec:res} 

We next evaluate \modelname{} on several captioning tasks, demonstrating state-of-the-art results. See supplementary for additional details.



\paragraph{Image Captioning.}
We compare \modelname{} caption quality to several baselines with different supervision levels, as presented in \cref{tab:captioning}(A). Here, all methods were trained end evaluated over the same dataset, using the commonly used MS-COCO \cite{lin2014microsoft, chen2015microsoft} and Flickr30k \cite{young2014image}. We begin by evaluating fully supervised techniques: BUTD \cite{anderson2018bottom}, UniVLP \cite{zhou2020unified}, ClipCap \cite{mokady2021clipcap}, Oscar \cite{li2020oscar}, and Lemon \cite{hu2022scaling}. As expected, these achieve a better score than \modelname{}, as they exploit the additional supervision of image-text pairs. Nevertheless, compared to the unsupervised approaches of MAGIC \cite{su2022language} and ZeroCap \cite{tewel2022zerocap}, \modelname{} achieves superior scores. Note that ZeroCap does not require any training data, while MAGIC requires text data similar to our setting.

\paragraph{Cross-Domain Captioning.}
We test our generalization ability by training on one dataset while evaluating on another, as in  \citet{su2022language}. Again, as can be seen in \cref{tab:captioning}(B), \modelname{} outperforms MAGIC \cite{su2022language}, which uses the same supervision as \modelname{}.

\begin{figure}
\centering
\includegraphics[width=1.0\textwidth]{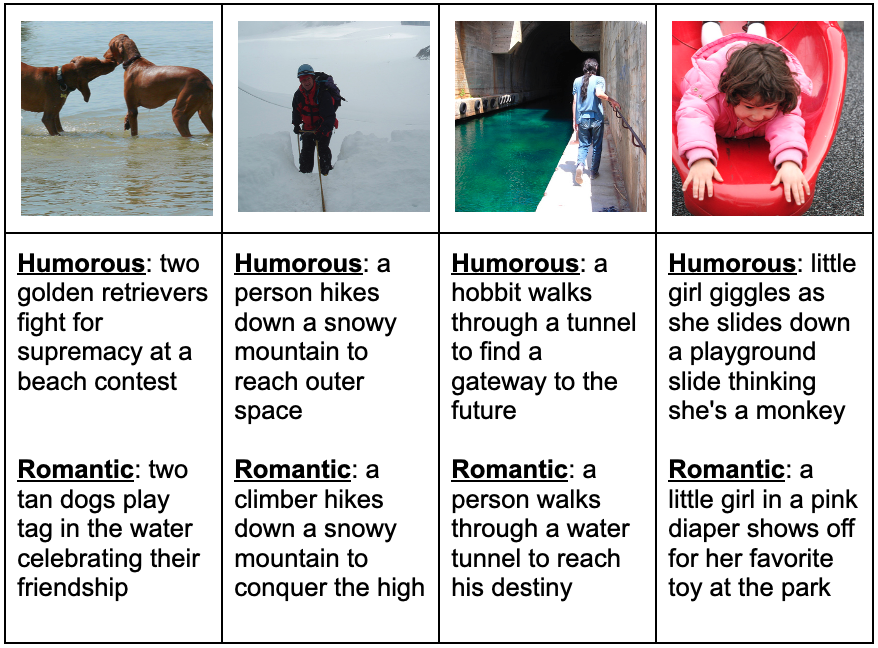}
\vspace{-0.35cm}
\caption{Example for styled captions of \modelname{} on FlickrStyle10K \cite{gan2017stylenet}.}
\vspace{-0.45cm}
\label{figures:examples.png}
\end{figure}

\paragraph{Style-Guided Captioning.}
Several works \cite{zhao2020memcap,gan2017stylenet} have studied the task of adapting a captioning model to a new style, such as ``romantic'' or ``humorous''. Since collecting paired examples for each style requires great effort, these consider the setting where the new style is only learned from text. This is easy to do in our setting, since we can train the decoder on any given style text. \cref{figures:examples.png} shows captions generated with \modelname{} in several styles (same setting and data as in \citet{zhao2020memcap}). \cref{tab:style} reports quantitative results for this setting, showing \modelname{} outperforms other baselines. To further analyze our approach, we present our results without pre-training (i.e., training on styled data only), with a text-only pre-training over COCO, and with text-image pre-training over COCO (similar to \cite{zhao2020memcap}). As can be seen, we outperform \cite{zhao2020memcap} even with considerably less supervision at pre-training. Moreover, both other variations improve results, demonstrating that CapDec can effectively use additional training data where available.

\begin{figure}
\centering
\includegraphics[width=1.0\textwidth]{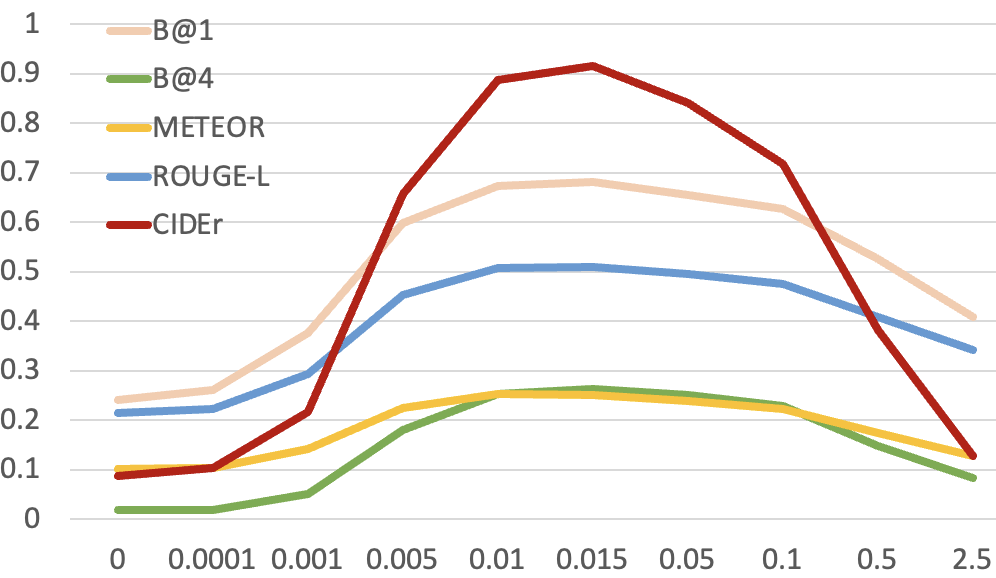}
\caption{\label{figures:varianceXmetrics.png}{The effect of the noise variance on MS-COCO performance.}}
\end{figure}

\paragraph{The Effect of Noise Level.}
A key element of \modelname{} is noise injection before decoding. 
To demonstrate the effect of noise, we report results as a function of the noise variance $\epsilon^2$ in \cref{figures:varianceXmetrics.png}. It can be seen that too little or too much noise is suboptimal. We note that the noise variance we chose, $\epsilon^2$=$0.016$,\footnote{As mentioned in \cref{sec:method}, we estimated the optimal STD by the mean infinity-norm of embedding differences between captions that correspond to the same image, which is $\epsilon$=$\sqrt{0.016}$} is based only on text, and not on the results in \cref{figures:varianceXmetrics.png} which are shown for analysis purposes only.

\begin{figure}
\centering
\includegraphics[width=1.0\textwidth]{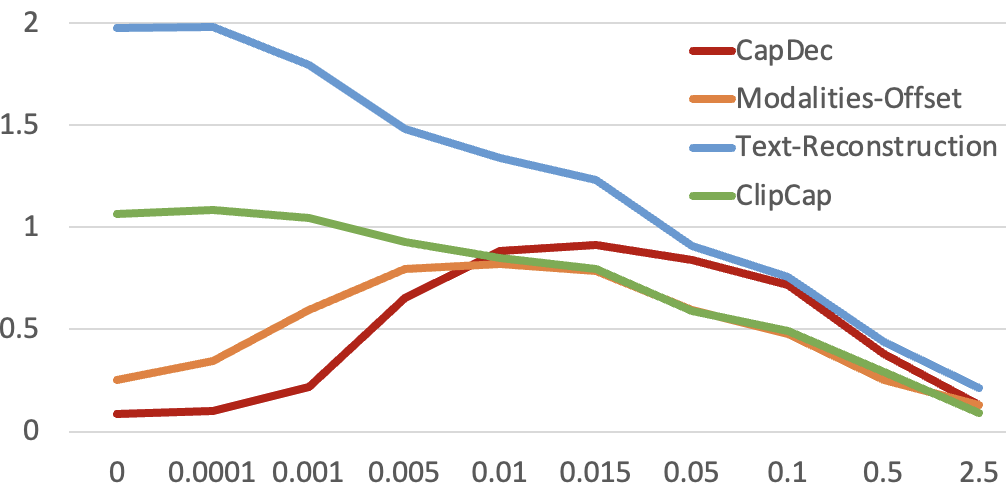}
\caption{\label{figures:noiseAblation.png}{
Analysis of performance of different methods as a function of the noise level (see Sec.\ref{sec:NoiseAblation}). We show the CiDER metric (higher is better), as
other metrics show similar trends. CapDec here is the same as in Fig.\ref{figures:varianceXmetrics.png}
}}
\end{figure}

\section{Noise Injection Analysis}
\label{sec:NoiseAblation} 
Noise-injection is a well-known technique for improving generalization \cite{reed1999neural, bishop1995training, an1996effects, vincent2010stacked}, and can be viewed as a data augmentation mechanism \cite{goodfellow2016deep}. In our case, the use of noise was also meant to address the modality-gap observed in \citet{liang2022mind}. In order to examine the specific effect of noise, we perform additional evaluations on COCO and show the results in Fig.\ref{figures:noiseAblation.png}. 

\paragraph{Text-Reconstruction:} We encode COCO captions using CLIP text embedding and decode them using the learned CapDec model. This does not involve images at all, and is meant to test whether noise injection simply serves as regularization for text auto-encoding. Fig.\ref{figures:noiseAblation.png} shows that adding noise does not help, and thus suggests that noise is not merely functioning as augmentation. 

\paragraph{ClipCap:} Recall that ClipCap is trained on joint text-image pairs \cite{mokady2021clipcap}. Here we trained ClipCap by adding noise to the image embeddings during training. It can be seen that noise does not improve performance,
 again suggesting that improvement is due to its specific role in domain-gap correction.

\paragraph{Modalities Offset:}  Given sufficient training paired-data, one could presumably learn the modalities-gap and correct for it. Here we test a simple approximation of the gap, that does not require image-text data to be paired, by calculating the shift between the mean of text embeddings and the mean of image embeddings in COCO. Then, given an image, we add the shift to its embedding to ``correct'' for this gap, and apply the CapDec trained decoder to the resulting embedding. Had this mapping been perfect, CapDec would not have needed additional noise injection. The results in Fig.\ref{figures:noiseAblation.png} show that the offset-correction does outperform CapDec at $\epsilon^2<0.01$, but underperforms overall. This suggests that the gap was not perfectly estimated, and that noise injection still serves to mitigate it. We leave it for future research to consider a more complex or fully-supervised model that learns the modality-gap explicitly.


\comment{
\paragraph{Fully-Supervised Settings}
In order to examine if the noise-injection is effective at CapDec just as a data-augmentation and not because of the modality-gap, we tried it also in fully-supervised settings. The ClipCap model \cite{mokady2021clipcap} shares the same high-level architecture as CapDec of encoder-decoder but it is trained with paired data in a fully-supervised manner. We injected noise into ClipCap at the latent space, before its decoder. For CapDec itself, we look at its performance for text-reconstruction, when the decoder's input is produced by CLIP text-decoder as done in its training. As can be seen in Fig.\ref{figures:noiseAblation.png}, for the two variations of the fully-supervised settings, any additional noise-level hurts performance. Thus, we conclude that the effectiveness of the noise-injection in CapDec, is highly related to the modality-gap, since otherwise it should be effective also for supervised settings. 

\paragraph{Modality Offset}
We tried to bridge the modality gap differently, by adding the modality-offset explicitly. We calculated the offset, in COCO, as the distance vector between the centers of the modalities, when for each modality, the center is calculated by averaging its CLIP embedding. Note that although the offset calculation requires image data, it still does not require it to be paired.  As can be seen in Fig.\ref{figures:noiseAblation.png}, adding the modality-offset, helps only in low noise-levels, and it anyway does not outperform the optimal noise level. It further support the claim that the noise-injection handles the modality-gap specifically.
}

\section{Conclusion}
The image captioning task has been extensively studied, with considerable progress in recent years. However, the number of available training datasets, containing image-text pairs, is still rather limited. Consequently, image captioning models inherit the limitations of their training data, such as biases \cite{hendricks2018women} or confinement to neutral style \cite{gan2017stylenet}. In this work, we suggest a new paradigm, where a generic vision-language model (e.g., CLIP) is adapted to image captioning using a text-only dataset. Furthermore, we demonstrate a simple and intuitive technique to overcome the inherent domain gap of CLIP \cite{liang2022mind}. For future work, we plan to study text-only training for other tasks, such as visual question answering and visual scene graph generation.

\section{Ethics Statement}
Image captioning models are notorious for their internal biases \cite{hendricks2018women}. These biases are usually inherited from the training data itself. We observe that since balancing a text-only dataset is much more feasible than collecting balanced text-image pairs, \modelname{} can be used to mitigate those biases. For instance, consider the problem of a dataset containing significantly more images of snowboarding men than women. Collecting more images requires substantial effort while replacing "man" with "woman" (and their synonyms) in all captions is quite simple. Therefore, our text-only training might mitigate some of the inherited bias.

\section{Limitations}
We observe that although CapDec achieves superior results compared to the baselines that use only text at training, it is still outperformed by fully supervised baselines. Since CLIP captures rich semantics in its latent space, we believe that text-only training can be further improved up to the almost same quality as supervised techniques in future work. In addition, note that \modelname{} relies on CLIP and a language model both of which were pre-trained on large English corpora. Therefore, we find the important task of extending \modelname{}'s capabilities to other languages to be a significant challenge.


\section*{Acknowledgments}
This work was supported by the Blavatnik Interdisciplinary Research Center (ICRC). We thank to Amir Hertz for sharing relevant code parts from his work on ClipCap \cite{mokady2021clipcap}.


\clearpage

\bibliography{egbib}
\bibliographystyle{acl_natbib}

\appendix
\section{Appendix}
\subsection{Implementation Details}
We use the RN-50x4 backbone for CLIP image encoder, and GPT-$2$ (large) as our language model (implementation of Wolf et al.\cite{wolf-etal-2020-transformers}). Following ClipCap \cite{mokady2021clipcap}, for the decoder architecture, we use a transformer-based \cite{vaswani2017attention} mapping network where we set the CLIP embedding length of $K=40$ with additional $K=40$ constants tokens and use $8$ multi-head self-attention layers with $8$ heads each. 
For optimization, we employed AdamW \cite{Kingma2015AdamAM} with weight decay as introduced by Loshchilov et al.~\cite{loshchilov2017decoupled}, with a learning rate of $2e^{-5}$ and $5000$ warm-up steps. 


\subsection{Datasets and Evaluation Metrics}
When evaluating over MS-COCO \cite{chen2015microsoft} and Flickr30k \cite{plummer2015flickr30k}, we followed Karpathy\cite{karpathy2015deep} split, similar to \cite{su2022language} and \cite{mokady2021clipcap}. Considering the FlickrStyle10K \cite{gan2017stylenet} dataset, we followed \cite{zhao2020memcap}, and split the  dataset randomly to $6/7$, and $1/7$ of training and test sets, correspondingly. For qualitative evaluation, we employ the commonly used BLEU \cite{papineni2002bleu} (B@1,B@4), METEOR \cite{denkowski2014meteor} (M), ROUGE-L \cite{lin2004automatic} (R-L), and CIDEr \cite{vedantam2015cider} (C) metrics.

\subsection{Qualitative Comparison}
All qualitative scores were reproduced or obtained from the works of \cite{su2022language} and \cite{zhao2020memcap} after carefully validating we use the same splits. Our metrics implementation is adapted from the official implementation of \cite{li2020oscar}.



\label{sec:appendix}

\end{document}